\pdfoutput=1
\documentclass[11pt]{article}

\usepackage[]{acl}

\usepackage{times}
\usepackage{latexsym}

\usepackage[T1]{fontenc}
\usepackage[utf8]{inputenc}

\usepackage{microtype}

\usepackage{hyperref}
\usepackage{enumitem}
\usepackage{booktabs}
\usepackage{multirow}
\usepackage{graphicx}
\usepackage{makecell}

\title{
From Rewriting to Remembering:\\
Common Ground for Conversational QA Models}

\author{
Marco Del Tredici, Xiaoyu Shen, Gianni Barlacchi, Bill Byrne, Adrià de Gispert\\
Amazon Alexa AI\\
  \texttt{mttredic|gyouu|gbarlac|willbyrn|agispert@amazon.com} \\}

\begin{document}
\maketitle
\begin{abstract}
In conversational QA, models have to leverage 
information in previous turns to answer upcoming questions. 
Current approaches, such as Question Rewriting, struggle to extract relevant information as the conversation unwinds.
We introduce the Common Ground (CG), 
an approach to accumulate conversational information as it emerges 
and select the relevant information at every turn.
We show that CG offers a more efficient and human-like way to exploit conversational information compared to existing approaches, 
%is more efficient than existing approaches in leveraging relevant information, 
leading to improvements on Open Domain Conversational QA.

\end{abstract}

\section{Introduction}
\label{sec:introduction}

Speakers involved in a conversation continuously share new information, and build on it to achieve their communicative goals. In human communication, this process takes place effortlessly.
As QA systems become conversational, efforts were made to make them able to mimic human behaviour, and to interpret the question at a turn in a conversation, based on the information in the previous turns. 
%collect and leverage the relevant information emerged in the interaction with the user. 
%
%All these methodologies aim at making QA models able to interpret the meaning of a question based on the contextual information previously emerged in the conversation.\m{probably rephrase needed}.
%
An approach to this task is to concatenate the previous turns to the current question \citep{christmann2019look,ju2019technical,qu2019attentive}. 
%The approach has two main shortcomings: it introduces a great amount of noise, since not everything in the previous turns is relevant; and, as the context grows, it is not manageable for Transformer models.
The approach has a main shortcoming, namely, it introduces a great amount of noise, since not everything in the previous turns is relevant.
%; and, as the context grows, it is not manageable for Transformer models.
%
An alternative approach is Question Rewriting (QR), in which the question is rewritten in a self-contained form based on the previous conversational information \citep{vakulenko2021question,anantha2020open}. 
QR selects only the relevant information in previous turns, thus improving over concatenation. 
However, as the conversation progresses and the amount of information grows, QR models often fail to compress it in a rewrite.
We argue that this is not only a limitation of the models, but an intrinsic limit of this approach, since producing informative rewrites is often unnatural also for humans (see Section \ref{sec:results_and_analysis}).
%the need to include all the previous information in an extended and well-formed version of the original question hinders the possibility to consider large amount of information \citep{deltredici2021questionrewriting}. 
%Furthermore, as generative QR systems take as input the full conversational context, they tend to consider the information emerged later in the conversation, while forgetting the earlier one \m{[ref needed]}.
% 

In this work, we address the shortcomings above.
Inspired by the studies of \citet{clark1996using},
%on how humans share information in conversations, 
we propose a methodology to represent conversational information
%the contextual information which emerges in interactions 
as a set of propositions, named the \textit{Common Ground} (CG): 
At each turn, the relevant information 
%shared by the speaker and the QA model 
is distilled in one or more propositions, which are added to the CG. 
As a new question comes in, the model selects the relevant information in the CG, and uses it to answer the question.
The CG can thus be considered as an \textit{optimized} summary, 
which returns the relevant information at every turn
%a dynamic object, which evolves based on what happens in the conversation, 
while keeping all the information discussed so far.

We use the QReCC dataset \citep{anantha2020open} to test CG on the task of Open-Domain Conversational QA (ODCQA) - 
in which answers to questions in a conversation have to be found in a large collection of documents - 
and show that it improves over existing approaches for modelling conversational information. 
%
%In our analysis, 
We show that this is due to the fact that CG implements a more efficient and human-like way to account for previous information, which takes the best of existing approaches while avoiding their shortcomings: 
%thanks to its ability to access the full conversational context, distill the relevant information in it, and make it available throughout the conversation, thus offering a more effective and human-like solution compared to existing approaches.
on the one hand, 
%similarly to concatenation, 
CG can access and maintain the full previous conversational context, but it avoids the noise issue;
%and size issues. 
on the other, 
%Similarly to QR, 
it can distill relevant information, but it is not forced to compress it in a single rewrite.

\section{Common Ground}
\label{sec:common_ground}

We now detail how we created a dataset for CG, and 
the model we implemented to generate the CG.

\subsection{Building the CG}
\label{subsec:building_the_cg}

We devise the CG as a set of propositions summarizing the information in a conversation.
%
% Out goal is to train a model to generate such set. 
%
%To the best of our knowledge, no dataset annotated for CG is available.
Since no dataset annotated for CG is available for QA, we created it.
We use QReCC \citep{anantha2020open}, a dataset for QR consisting in a set of conversations. For each turn in a conversation, the original question \textit{q} and its rewrite \textit{r} are provided. 
Intuitively, the rewrite makes explicit the entities discussed in the conversation. 
If \textit{q} is self-contained, then \textit{q}=\textit{r}.
We define a proposition in the CG as any sequence of words in the rewrite which are nouns, adjectives or entities.\footnote{Identified using Spacy: \url{https://spacy.io/}.}
For example, given \textit{$q_1$} `how old is Messi?', the rewrite \textit{$r_1$} is equal to \textit{$q_1$}, and \textit{$CG_1$} is \{`\textit{Messi}'\}. Given \textit{$q_2$} `which position does he play?', \textit{$r_2$} is `which position does Messi play?' and \textit{$CG_2$} is \{`\textit{Messi}', `\textit{position}'\}.
We use this approach to enrich each turn in QReCC with the gold CG. 
%create the gold CG for each turn, in each conversation in QReCC.
%
%While we are aware that our to create the gold CG includes some arbitrary choices, and other approaches might be equally valid, we believe it is a reasonable staring point.

Importantly, $\sim$70\% of the conversations in QReCC were collected by showing the speaker the title and first sentence of a Wikipedia article \citep{anantha2020open}. This information is often crucial to understand a question, especially at turn 1 (e.g., title: `Albert Camus', $q_1$: `When was he born?'), but, potentially, also at subsequent turns ($q_2$: `What did he write?').
We therefore collect the relevant Wikipedia information (which we call $doc$), and use it to further enrich QReCC conversations.\footnote{The details about the enriched dataset are in Appendix \ref{sec:enriching_qrecc}.}
%pass it to the model together with the conversational context. 
Note that $doc$ is the same at every turn in the conversation. 
We refer to the union of conversational and Wikipedia information as \textit{contextual} information.
Finally, since QReCC only includes train and test split, we randomly sample 20\% of the train and use it as validation set.
%Ultimately, the input for the Generator at turn \textit{n} is $doc \mathbin\Vert conv_{[0:n-1]} \mathbin\Vert q_n$. We refer to the union of $doc$ and $conv_{[0:n-1]}$ as \textit{contextual} information.

\subsection{Predicting the CG}
\label{subsec:predicting_the_cg}

We introduce a model to produce the CG, which consists of two modules: \textit{Generator} and  \textit{Selector}. 

\paragraph{Generator} At turn $t_n$, the Generator is trained to generate the gold CG $CG_n$ given $doc \mathbin\Vert conv_{[0:n-1]} \mathbin\Vert q_n$,
where $\mathbin\Vert$ indicates concatenation, 
$doc$ is the information from Wikipedia,
$conv_{[0:n-1]}$ is the concatenation of questions and answers from turn $t_0$ to $t_{n-1}$, 
and $q_n$ is the current question.
%
%Importantly, $\sim$70\% of the conversations in QReCC were collected by showing the speaker the title and first sentence of a Wikipedia article \citep{anantha2020open}. This information is often crucial to understand a question, especially at turn 1 (e.g., title: `Albert Camus', $q_1$: `When was he born?'), but, potentially, also at subsequent turns ($q_2$: `What did he write?').
%We therefore collect from Wikipedia the relevant information (which we call $doc$), and pass it to the model together with the conversational context. Note that $doc$ is the same at every turn in the conversation.\footnote{The details about the enriched dataset are in Appendix \ref{sec:enriching_qrecc}.} 
%
%Ultimately, the input for the Generator at turn \textit{n} is $doc \mathbin\Vert conv_{[0:n-1]} \mathbin\Vert q_n$. 
%We refer to the union of $doc$ and $conv_{[0:n-1]}$ as \textit{contextual} information.
%
We implement the Generator using a T5-base model.\footnote{The details of Generator and Selector are in Appendix \ref{sec:model_for_cg_prediction}.}
%We finetune the parameters on the validation set.
We train the generator using the enriched QReCC.
%Since QReCC only includes train and test split, we randomly sample 20\% of the train and use it as validation set.
%

\paragraph{Selector} The propositions returned by the Generator for every turn are stacked in the CG. However, as the conversion goes on, some of the propositions are no longer relevant. The role of the Selector is to select only the relevant propositions in the CG. 

\begin{figure}
  \includegraphics[width=\linewidth]{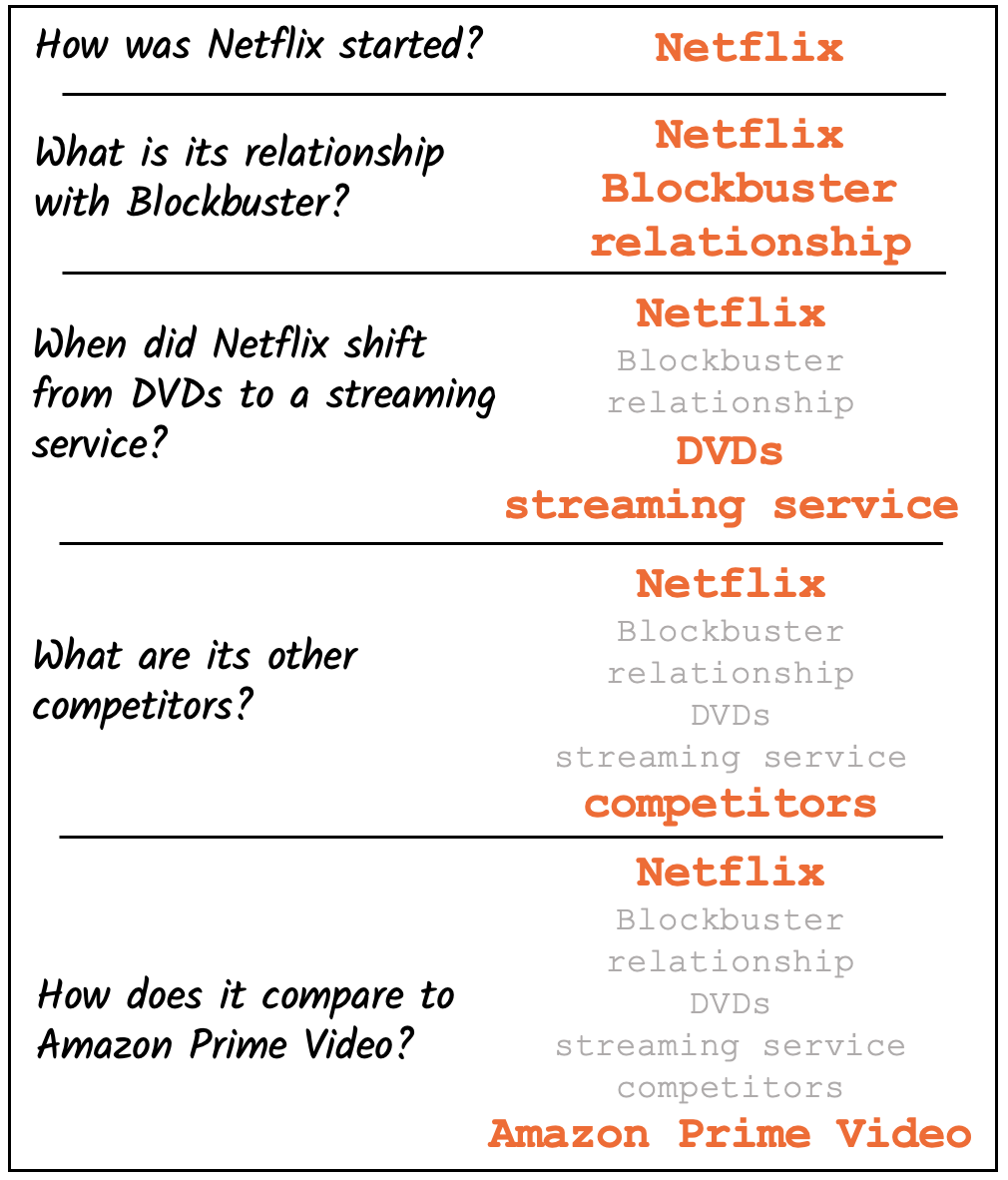}
\caption{On the left, the questions from the user; on the right, the CG generated by the Generator: 
highlighted the propositions selected by the Selector at each turn, in grey those kept in the CG but not selected.}
\label{fig:cg}
\end{figure}

We implement the Selector as a binary classifier. 
To create the data to train the model, we use again QReCC: given the full CG available at turn $n$, we label as 1 the propositions in it that occur in the gold answer span, 0 otherwise. The rationale behind this approach is: an item in the CG is relevant if it is mentioned in the answer. We train the model on the QReCC train split.
At test time, we label the propositions in the CG, and keep only those labelled as 1.
Figure \ref{fig:cg} shows an example of CG.

\section{Experiments}
\label{sec:experiments}

The goal of accounting for contextual information is to improve the performance on a downstream task.
%All the methodologies that enable QA models to account for conversational information share the same goal, namely, improve the performance on a downstream task.
Hence, we compare CG to existing approaches on the task of ODCQA.

\noindent \textbf{Data}
We use again QReCC, as it
%
% The dataset 
meets the requirements of the task: it is conversational, and it allows to experiment in an Open-Domain scenario.

\noindent \textbf{Pipeline}
%\paragraph{Pipeline}
We use a retriever-reader pipeline.
%, a common choice for the task of ODCQA [ref]. 
%
The retriever returns the top \textit{n} most relevant candidates from the set of documents; these are passed to the reader, which extracts the final answer. 
We use BERTserini \citep{yang2019end}, using BM25 as a retriever and a BERT-Large as a reader. 
Each candidate returned by the retriever has a score $s_{ret}$; the answer extracted from that candidate by the reader has a score $s_{rea}$. The final score $s$ for the answer is defined as: $(1 - \mu)\cdot s_{ret} + \mu \cdot s_{rea}$. 

For the retriever, we set \textit{n} to 20, and we follow \citet{anantha2020open} in setting $k_1$=0.82 and $b$=0.68. We tune the value of $\mu$ on the validation set independently for each approach (see Section \ref{sec:setups}).
We do not finetune the reader, as we want to assess how much the CG can directly benefit any QA model, without the need to finetune it.

\subsection{Setups}
\label{sec:setups}

%\ev{The pipeline defined above takes as input a question and returns an answer. 
%In our experiments, we assess how the performance changes when, besides the question, we provide additional information. 
%at turn $n$, the model is fed with additional contextual information extracted from Wikipedia ($doc_n$) and the previous turns in the conversation ($conv_{[0:n-1]}$).
%
%More specifically, }
We test the pipeline's performance when provided, at turn $n$, with each of the following inputs: 
\\
\textbf{original}: the original question $q_n$.
\\
\textbf{concat.}: the concatenation $doc \mathbin\Vert conv_{n-1} \mathbin\Vert q_n$.\footnote{Note that we use $conv_{n-1}$, and not $conv_{[0:n-1]}$, due to the 
max length limit of the reader of BERTserini.}
\\
\textbf{rewrite}: the rewrite $r_n$ produced with a T5-base model.
The model generates the rewrite based on $doc \mathbin\Vert conv_{[0:n-1]} \mathbin\Vert q_n$. 
\\
\textbf{summary}: the concatenation $summ_{[0:n-1]}\mathbin\Vert q_n$, where $summ_{[0:n-1]}$ is the summary of $doc \mathbin\Vert conv_{[0:n-1]}$, created with a T5-base model pretrained for summarization \citep{raffel2019exploring}.\footnote{The details of the Rewrite and Summarization models are in Appendix \ref{sec:generative_models_for_qr_and_summarization}.}
%\footnote{\url{https://ai.googleblog.com/2020/02/exploring-transfer-learning-with-t5.html}}
\\
\textbf{CG}: The CG predicted using our approach, concatenated with the current question: $CG_n \mathbin\Vert q_n$. 
\\
\textbf{CG-full}: The full CG generated up to turn $n$, i.e., we do not use the Selector module: $CG_n$-$full \mathbin\Vert q_n$. 
%Moreover, we will experiment with the same two version defined for CG-gold, \textbf{CG-full} and \textbf{CG-last}.

%\textbf{rewrite}: the rewrite $r_n$ produced by humans, from the the QReCC dataset. Recall that rewrites were created considering both the $doc_n$ and $conv_{[0:n-1]}$ (see Section [n])
%\\
%\textbf{CG-gold}: The gold CG as defined in Section \ref{subsec:building_the_cg}, concatenated with the current question. The gold CG is used to set the upperbound performance of our approach.  
%We experiment with two versions: \textbf{CG-gold-full}, in which \textit{all} the propositions in the CG are used:  $CGgold_{[0:n-1]} \mathbin\Vert q_n$; and \textbf{CG-gold-last}, in which only the Cg of the last turn is used: $CGgold_{n-1} \mathbin\Vert q_n$.

%\ev{Note that we make available, to any of the approaches defined above, the same contextual information, namely, the one derived from the Wikipedia page ($doc$) and the one from the conversation ($conv$). This enables a fair comparison of the different approaches.}
%
%\ev{Finally, in our experiments we follow \citet{deltredici2021questionrewriting}, who show that combining different approaches at the retriever and reader is beneficial for the final performance on the downstream task.}

\section{Results and Analysis}
\label{sec:results_and_analysis}
We show the results of our experiments in Table \ref{tab:main_results}.
We measure the performance on the target task in terms of F1, and use MRR and Recall@10/20 to assess the performance of the retriever.\footnote{We use the code by QReCC authors: \url{github.com/apple/ml-qrecc/tree/main/utils}.}
We also report the results obtained with gold (-g) rewrites and CG, where the latter is defined, at turn \textit{n}, as gold $CG$-$full_n$ for the retriever and gold $CG_n$ for the reader - i.e., the best combination observed in our experiments (see below).

\begin{table}[t]
\begin{center}
\begin{tabular}{l|c|c|c|c}
\toprule
\textbf{Approach} & \textbf{F1} & \textbf{MRR} & \textbf{R@10} & \textbf{R@20}\\
\midrule
original   & 6.23  & 2.89  & 5.56  & 6.65  \\
concat.    & 8.95  & 21.67 & 37.55 & 41.51 \\
rewrite    & 12.46 & 13.73 & 24.52 & 28.6  \\
summary    & 12.02 & 21.81 & 34.72 & 38.33 \\
CG         & 13.41 & 15.66 & 27.67 & 32.09 \\
CG-full    & 12.18 & 16.52 & 29.47 & 34.06 \\
CG-full/CG & 14.2 & 16.52 & 29.47 & 34.06 \\
\midrule
rewrite-g & 13.42 & 17.16 & 29.07 & 33.26 \\
CG-g & 15.17 & 17.95 & 31.18 & 35.65 \\
\bottomrule
\end{tabular}
\caption{Results on the QReCC test set. CG-full/CG indicates that we used CG-full for the retriever and CG for the reader.}
\label{tab:main_results}
\end{center}
\end{table}
%
%In line with previous work, our results show that 
As expected, approaches leveraging contextual information improve over the original question. 
Among these approaches, CG is the best: 
%
%In particular, CG 
it improves the performance over rewrite, and, remarkably, it matches the results obtained with \textit{gold} rewrites.
A further improvement in F1 is observed when using CG-full at the retriever and CG at the reader (CG-full/CG), while using only CG-full degrades the performance. 
%
%These results are consistent with those in \citet{deltredici2021questionrewriting}, and indicate that CG-full, which is more informative but more noisy, is the best option for the BM25 retriever, which is robust to noise. Conversely, the BERT reader performs better when fed with the filtered information in CG. 
%
This shows that using the more informative but potentially noisier CG-full improves retrieval, but one needs to feed the filtered information from CG to the reader to see improvements in F1, as also observed by \citet{deltredici2021questionrewriting}.
The different response to noise also explains the results of concatenation,
%and, in part, summaries (see below), 
which obtain high performance in retrieval, but drops in F1.

\paragraph{CG vs. QR}
In Table \ref{tab:examples}, we show examples from QR and CG. 
\begin{table*}[t]
\begin{center}
\begin{tabular}{l|c|c|c}
\toprule
 & \textbf{Original Question} & \textbf{Question Rewriting} & \textbf{Common Ground}\\
\midrule
 1 & 
 \small{\makecell{What's the average\\starting salary in the UK?} } & 
 \small{\makecell{What's the average starting salary\\for a physician assistant in the UK?}} & 
 \small{\makecell{\{\textit{the average starting salary}, \\ \textit{the UK}, \textit{a physician assistant} \}}} \\
 \midrule
 2 & 
 \small{\makecell{What about in the US?} } & 
 \small{\makecell{What about in the US?}} & 
 \small{\makecell{\{\textit{the average starting salary}, \\ \textit{the US}, \textit{a physician assistant} \}}} \\
 \midrule
 3 & 
 \small{\makecell{Are flows bidirectional?} } & 
 \small{\makecell{\underline{Are flows bidirectional?}}} & 
 \small{\makecell{\{\textit{data  
 network architectures}, \textit{edge switches}, \\ \textit{bidirectional flows},  \textit{FAT tree topology}, \\ \textit{upstream packet}, \textit{routes}, \textit{core}, \textit{aggregator}\}}} \\
 %\midrule
 %4 & 
 %\small{\makecell{When was it brought\\to the British isles?} } & 
 %\small{\makecell{When were scrapers brought\\to the British isles?}} & 
 %\small{\makecell{\{\textit{the British isles}, \\ \textit{the neolithic revolution}\}}} \\
 %
\bottomrule
\end{tabular}
\caption{Examples of rewrites and CG. Predicted rewrites are in plain text, gold rewrites underlined. }
\label{tab:examples}
\end{center}
\end{table*}
In row 1, both approaches extract the relevant information from the previous turns - in a conversation about physician assistants. 
In the next turn (2), QR fails to expand the question and to substitute
`about' with the contextual information, due to the large amount of information required (`the average starting salary for a physician's assistant in the US'). 
We often observe this limitation for the QR model.
%QR often shows this limitation. 
This is not the case for CG, since here the information grows \textit{incrementally}, i.e., the information from the current turn (`the US') is added \textit{on top} of the one already present, while non relevant information
(`the UK') is discarded.
% 
%While, in some cases, models are not able to produce a rewrite, in others this is just not possible.

In the previous case, the QR model fails to produce a rewrite; in others, this is just not possible. 
%the previous example shows a limitation of the model in producing a rewrite, others show that, sometimes, this is just not possible. 
In the 6th turn of a conversation about different kinds of data network architectures (row 3), the user asks a general question about flaw types which encompasses all the previous information: 
%In this case, 
there is so much information to compress, here, that \textit{not even} humans manage to do it, and the gold rewrite is the same as the original question.\footnote{We provide in Appendix \ref{sec:example_of_conversation} the whole conversation, plus additional examples of (nearly) impossible rewrites.} 
CG sidesteps this problem simply by making available all the pieces of relevant information emerged in the conversation, which can be selected and exploited by the model,
%thanks to the incremental accumulation of the relevant information
%, as it makes available the 
%maintains the main 
%information 
%from previous turns, 
%which is readily available, 
without the need to produce a long natural sentence. 
Note that besides being more effective, this solution is also more human-like: 
Speakers do not \textit{repeat} all the contextual information as they make a question, but, rather, they \textit{remember} the key points of the conversation. 
%CG offers a more easy and human-like solution to sidestep this problem, as the relevant contextual information is distilled and kept after each turn. 
%the recency bias mentioned in Section \ref{sec:introduction}.

%Finally, in a conversation about the Neolithic Revolution, `scrapers' are mentioned at turn 5. In the following turn, the user asks about again about the main topic, i.e., the Neolithic Revolution (row 4), but, due to the recency bias, the QR model rewrites the question resolving the reference incorrectly. Differently, in CG the correct information is readily available, and picked by the Selector.

\paragraph{CG vs. Summary}
Summaries convey all contextual information, which makes them suitable for the retriever, but not for the reader.
%The summarization approach yields high results in retrieval, but it does not match QR and CG on the downstream task. 
%We believe that this is due to the summaries being too general, as they convey all the contextual information. 
CG is superior because, as said above,
%as it can be considered as 
is an \textit{optimized} summary conditioned on the current question.
%including only relevant information, identified based on the current question. 
%
In fact, when we create the CG without considering the current question, the model cannot identify the relevant information, and the results are comparable to those of summary (F1=12.6).
%, as the the relevant information cannot be identified. 
%
For example, for the question `where did he come from?', the CG predicted in the normal scenario is \{\textit{Rick Barry}\}, while, without the current question, is \{\textit{the ABA}, \textit{free-throw percentage}, \textit{the 1968–69 season}, \textit{Rick Barry}\}.

\paragraph{\textit{Conv} vs. \textit{Doc}}
We measure the performance for the best setup (CG-full/CG) when the CG is created considering either $doc$ or $conv$: with the former, the F1 is 13.38, with the latter 13.65.
The decrease in performance of $doc$ and $conv$
%fact that, in both cases, the performance decreases 
compared to $doc$+$conv$ indicates that considering multiple source of information is beneficial for the overall performance of the model.
Also, the fact that $conv$ yields better results than $doc$ is expected: in QReCC, the information from $doc$ is mostly leveraged at the first turn, while the information from $conv$ is relevant throughout the full conversation. 

\section{Related Work}

Approaches to modelling conversational information have used either sparse or dense representation \citep{qu2019history,qu2019attentive,Qu2020}. 
This work focuses on the former.
In this group, 
concatenation was proposed as an initial approach \citep{christmann2019look,ju2019technical,qu2019attentive}, followed by
Question Rewriting \citep{elgohary2019}. 
The main models for QR are either generative \citep{vakulenko2021question,yu2020few} or extractive one \citep{Voskarides2020} - i.e., the relevant tokens in the context are appended to the question.
When a single model is used for both retriever and reader, generative model overperform extractive ones \citep{vakulenko2021comparison}; however, mixing the two approaches further improves the performance \citep{deltredici2021questionrewriting}.
Our work is related to \citep{Voskarides2020}, as we also aim at extracting the relevant contextual information. However, instead of appending this information to the question, we stack it in the CG, and enable the model to pick the relevant information at each turn.

%generative  vakulenko2021question, yu2020few
%expansive Voskarides2020
%comparison vakulenko2021comparison, deltredic

\section{Conclusions}

We introduced the Common Ground, a novel approach for leveraging contextual information.
We show that CG outperforms the main existing approaches in the ODCQA task, due to its ability to select and maintain the relevant information in a more effective and human-like way.

We see two main directions for future research on CG.
First, we will exploit the ability of CG to include several kinds of information to make it more informative.
%, including additional information.
For example, to answer the question `how many Covid cases today?', a QA system needs to be aware of the \textit{time} and \textit{location} of the person asking it \citep{zhang2021situatedqa}. We want to include these and other information in the CG.
Second, we want to use CG to make QA models more transparent.
Currently, virtual assistants (such as Alexa, Siri and Google Assistant) are black boxes, i.e, the user does not know which information they extract from the input question, and which one they leverage to provide answers. 
This can make the interaction with them frustrating. 
%hinders the possibility of a successful and frustration-free interaction. 
CG offers a solution to the problem, as it allows to see what the assistant \textit{has in mind} at each conversational turn. 
We will conduct experiments in which the CG is shared with the user, and see how this can make the interaction with the assistant more engaging and successful.

\bibliography{anthology,custom}
\bibliographystyle{acl_natbib}

\appendix

\section{Enriching QReCC}
\label{sec:enriching_qrecc}

Approx. 78\% of the conversations in QReCC are derived from the QuAC dataset (\url{https://quac.ai/}). In QuAC, dialogues are created by showing to the student (i.e., the person making questions) the title of the section of a Wikipedia page and the first sentence of first paragraph in the page. We retrieve this information from the QuAC dataset, and add it to the QReCC dataset. 
As mentioned in the main paper, we add the information from Wikipedia to all the turns in a conversations. As a results, 76.5\% of the datapoints in the train split and 71.8\% of those in the test split have additional information. 
We will release the code for enriching QReCC with CG and Wikipedia information upon publication.

\section{Model for CG prediction}
\label{sec:model_for_cg_prediction}

\paragraph{Generator}
In order to generate the CG, we use the T5-base model available at: \url{https://huggingface.co/transformers/model_doc/t5.html}.

We fine-tuned the model on the task of generating the CG with the following parameters: 
max source length= 512;
max target length= 64;
val max target length= 64;
evaluation strategy= steps;
num train epochs= 5;
per device train batch size= 4;
per device eval batch size= 8;
eval steps= 82;
seed= 42;
warmup steps= 500;
eval beams= 5;
learning rate= 5e-5.

\paragraph{Selector}
In order to select the relevant propositions in the CG, we use the DistilBert model available at: \url{https://huggingface.co/transformers/model_doc/distilbert.html}.

We fine-tuned the model with the following parameters: 
max source length= 512;
evaluation strategy= steps;
num train epochs= 5;
per device train batch size= 16;
per device eval batch size= 64;
eval steps= 82;
seed= 42;
warmup steps= 0;
learning rate= 5e-5.\\

%We will release the code of our model upon publication. 

\section{Generative models for QR and Summarization}
\label{sec:generative_models_for_qr_and_summarization}

\paragraph{QR model}
In order to generate the rewrites, we use the same T5-base model used to implement the Generator. We fine-tuned the model on the QR task using the QReCC train split, with the same parameters reported in Appendix \ref{sec:model_for_cg_prediction}. 
\paragraph{Summarization model}
In order to generate the summaries, we use again the same T5-base model used for the Generator and the QR model. In this case, however, we do not need to fine-tuned the model, since it was already optimized for the task: to generate the summaries, we simply provide to the model as input the string `summarize: ' followed by the contextual information.

\section{Example of conversation}
\label{sec:example_of_conversation}

We report below the full conversation up to the question used as an example in Table \ref{tab:examples}, row 3.\newline
\break
\noindent $q_1$: What are scalable data center network architectures?\\
$a_1$: DCNs need to be scalable and efficient to connect tens or even hundreds of thousands of servers to handle the growing demands of Cloud computing.\\
$q_2$: What are some examples of scalable data center network architectures?\\
$a_2$: 1 Three-tier DCN 2 Fat tree DCN 3 DCell\\
$q_3$: Describe the characteristics of FAT tree topology\\
$a_3$: In a fat tree, branches nearer the top of the hierarchy are fatter (thicker) than branches further down the hierarchy. In a telecommunications network, the branches are data links; the varied thickness (bandwidth) of the data links allows for more efficient and technology-specific use.\\
$q_4$: What routes can be taken by an upstream packet?\\
$a_4$: The router is upstream of the computer, connecting the computer to the whole internet. ... Each router does not need to know the whole route to the destination;\\
$q_5$: Describe core, aggregator and edge switches.\\
$a_5$: In small networks of a few hundred users, edge switches can be connected redundantly directly to core switch/router devices. However, for larger networks, , an additional layer of switching, called the distribution layer, aggregates the edge switches.\\

In Table \ref{tab:examples_2}, we report examples for which the gold rewrite provided in the QReCC dataset is equal to the original question, despite the fact that the question needs contextual information to be correctly understood. For each example, we provide the information in the CG, and a comment about why creating a rewrite is not possible, or very unnatural. Due to space reasons, we do not report the full conversation. However, we report the conversation and turns IDs, which can be used to look up for the full conversation in the QReCC dataset available at \url{https://github.com/apple/ml-qrecc/tree/main/dataset}.

\begin{table*}[t]
%\begin{center}
\begin{tabular}{l|l}
%\toprule
%\textbf{id} & \textbf{Question} & \textbf{QR} & \textbf{CG}\\
\midrule
17-10 &  \makecell[l]{\textbf{Question}: What form of energy is used in eating?                  \\ 
                \textbf{Common Ground}: \textit{energy, light energy, heat energy, gravitational energy, form, type,}\\\textit{motion, mechanical energy, examples, potential energy, electrical energy, sound energy,}\\ \textit{chemical energy, nuclear energy, atomic energy, kinetic energy} 
                \\ 
                \textbf{Comment}: the question comes at the end of a long conversation, and refers to the\\previously mentioned forms of energy. The hypothetical QR should include them all:\\\textit{What form of energy, among light energy, heat energy, [..] is used in eating?} 
                }
\\
\midrule
22-9 &  \makecell[l]{\textbf{Question}: What is the oldest spice?                  \\ 
                \textbf{Common Ground}: \textit{spices, cumin, world, coriander, cilantro, herb, garlic, oregano, }\\\textit{root, stem, seed, fruit, flower, bark, tree, plant, Indian, pepper, Nutmeg, mace, Mustard,}\\ \textit{seeds, Fenugreek, Turmeric, Saffron} 
                \\ 
                \textbf{Comment}: similarly to the previous example, the question comes at the end of a \\long conversation, and refers to all previous information. The hypothetical QR should be: \\\textit{What is the oldest spice among cumin, coriander [...]?} 
                }
\\
\midrule
28-4 &  \makecell[l]{\textbf{Question}: What can I do as an individual level?   \\ 
                \textbf{Common Ground}: \textit{global warming, long-term rise, average temperature,}\\\textit{Earth's climate system, climate change, temperature measurements, dangers, scientists,}\\ \textit{ sea ice, sea level rise, heat waves, methods, Carbon dioxide, oil, coal, fossil fuels, energy,}\\\textit{homes, cars, smartphones} 
                \\ 
                \textbf{Comment}: again, the user's question encompasses all previous conversation,\\ in which several problems related to global warming were mentioned. A (tentative) rewrite\\which captures the information up to this point should therefore be of the kind:\\\textit{What can I do in order to better use energy for my home, car, smartphone, thus reducing}\\\textit{the emission of carbon dioxide and reduce impact on global warming?} 
                }
\\
\midrule
583-6 &  \makecell[l]{\textbf{Question}: Was there anyone opposed to him in this?   \\
                \textbf{Common Ground}: \textit{Ira Hayes, World War II, civilian life, war, family, 1946,}\\\textit{Gila River Indian Community, Edward Harlon Block, Hank Hansen, flag-raiser}\\\textit{controversy, Marine Corps} 
                \\ 
                \textbf{Comment}: in this dialogue, many facts about Ira Hayes are explained. The original\\question refers to several of them, and a (very tentative) rewrite should be like:\\\textit{Was there anyone opposed to Ira Hayes in revealing the truth that Harlon Block was still}\\\textit{being misrepresented publicly as Hank Hansen?}
                }
\\
\midrule
590-6 &  \makecell[l]{\textbf{Question}: What was the impact of this column?   \\
                \textbf{Common Ground}: \textit{Israel, Krauthammer, Oslo accords, 2006 Lebanon War, column,}\\\textit{Let Israel Win the War} 
                \\ 
                \textbf{Comment}: also in this case, the conversation touches upon several related facts,\\and in order to correctly interpret the question in the light of such facts,\\it should be rewritten like:\\\textit{What was the impact of the column 'Let Israel Win the War' written by Krauthammer}\\\textit{during the 2006 Lebanon War, in which he opposes the Oslo accords?}
                }
\\
\bottomrule
\end{tabular}
\caption{Examples in which the rewrite is nearly impossible or very unnatural. In the left column we report the conversation-turn IDs.}
\label{tab:examples_2}
%\end{center}
\end{table*}

\end{document}